\documentclass[10pt,twocolumn,letterpaper]{article}

\usepackage[pagenumbers]{cvpr} 

\newcommand{\mName}{EviPrompt }
\usepackage{pifont}
\usepackage{multirow}
\usepackage[dvipsnames]{xcolor}

\definecolor{cvprblue}{rgb}{0.21,0.49,0.74}
\usepackage[pagebackref,breaklinks,colorlinks,citecolor=cvprblue]{hyperref}

\def\paperID{*****} 
\def\confName{CVPR}
\def\confYear{2024}

\title{EviPrompt: A Training-Free Evidential Prompt Generation Method\\ for Segment Anything Model in Medical Images}

\author{Yinsong Xu$^{1,3}$, Jiaqi Tang$^{2,3}$, Aidong Men$^{1,}$, Qingchao Chen$^{3}$\thanks{Qingchao Chen is the corresponding author} \\
$^{1}$Beijing University of Posts and Telecommunications\\
 $^{2}$The Chinese University of Hong Kong\\
  $^{3}$National Institute of Health Data Science, Peking University}

\begin{document}
\maketitle

\begin{abstract}
Medical image segmentation has immense clinical applicability but remains a challenge despite advancements in deep learning. The Segment Anything Model (SAM) exhibits potential in this field, yet the requirement for expertise intervention and the domain gap between natural and medical images poses significant obstacles. This paper introduces a novel training-free evidential prompt generation method named \mName to overcome these issues. The proposed method, built on the inherent similarities within medical images, requires only a \textit{single} reference image-annotation pair, making it a \textit{training-free} solution that significantly reduces the need for extensive labeling and computational resources. First, to automatically generates prompts for SAM in medical images, we introduce an evidential method based on uncertainty estimation without the interaction of clinical experts. Then, we incorporate the human prior into the prompts, which is vital for alleviating the domain gap between natural and medical images and enhancing the applicability and usefulness of SAM in medical scenarios. \mName represents an efficient and robust approach to medical image segmentation, with evaluations across a broad range of tasks and modalities confirming its efficacy.
\end{abstract}

    

\section{Introduction}
Medical image segmentation plays a pivotal role across a wide variety of clinical applications, including diagnosis, and treatment planning. The recent surge in deep learning illustrates the significant potential for enhancing the capabilities of medical image segmentation. However, it remains a challenging task, continuing to attract the attention of a substantial number of researchers \cite{hu2023label, bai2023bidirectional,feng2023unsupervised}.

\begin{figure}[]
    \centering
    \includegraphics[width=0.85\linewidth]{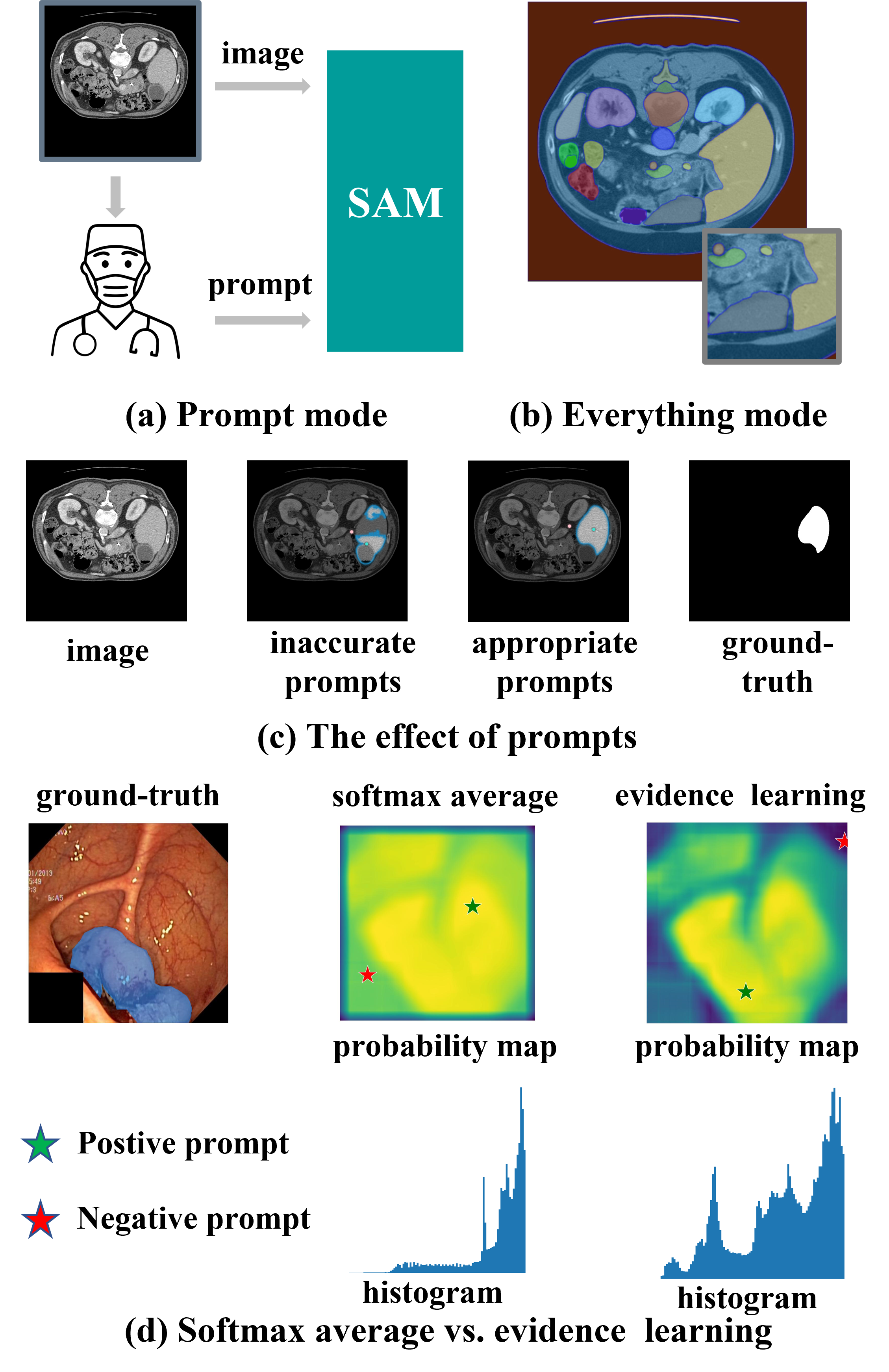}
    \caption{(a) Prompt mode requires interactions of clinical experts to provide prompts during the inference process. (b) The output in everything mode lacks associated semantic labels. Experts must discern or combine multiple masks to pinpoint the desired target. The liver is predicted as two masks. (c) Prompts directly influence the segmentation performance, and an imprecise prompt may compromise results, and an inaccurate prompt may compromise results. (d) The softmax output can lead to overconfidence, even for erroneous predictions. }
    \label{fig:teaser}
\end{figure}

The Segment Anything Model (SAM) \cite{kirillov2023segany} has demonstrated promising results in generating a diverse range of fine-grained segmentation masks. This methodology framework promotes a revolution in segmentation models and illuminates a path for future advancements. \textit{However, its application to medical images poses two significant challenges, elaborated as follows:}

\noindent\textbf{The requirement of expertise intervention}. SAM offers two modes of segmentation: the prompt mode and the everything mode. The former requires prompts (points, boxes, or masks) as input for each image during inference, as shown in Fig.\ref{fig:teaser}(a). Prompts play an essential role in directly indicating the Regions of Interest (ROIs). It presents a dilemma: obtaining precise prompts is difficult without professional medical knowledge or a pre-existing segmentation mask. Thus, interaction with clinical experts is necessary. In the latter mode, while SAM automatically generates potential masks, it lacks associated semantic labels(\eg assignment of liver class to a segmented ROI). Given that clinical applications often emphasize specific anatomical ROIs, experts must discern or combine multiple masks to pinpoint the desired target, as shown in Fig.\ref{fig:teaser}(b). In essence, both modes entail human intervention.

\noindent\textbf{The domain gap between natural and medical images}. SAM is trained on over one billion masks from 11 million images. The large-scale dataset enables SAM a strong generalization capability. Yet, recent studies \cite{he2023accuracy, mazurowski2023segment} indicate its unstable performances across diverse medical datasets. This instability is mainly because the training samples for SAM are mainly natural images. Transferring knowledge from natural images to medical tasks brings little gain due to the intrinsic differences in features and task specifications \cite{raghu2019transfusion}.

Prompts, as the main difference between SAM and existing segmentation methods, directly affect the segmentation performance \cite{mazurowski2023segment}. As illustrated in Fig.\ref{fig:teaser}(c), an inaccurate prompt may produce degraded segmentation results. It is also demonstrated in \cite{ji2023segment} that substantial human prior knowledge is helpful to alleviate the gap and obtain relatively accurate results. Inspired by it, a path appears to be paved by incorporating the human prior knowledge in prompts. Consequently, we utilize the prompt mode, and the solutions to the previous two challenges lead to the following two fundamental questions: \textit{Q1: How to automatically generate prompts?} \textit{Q2: How to inject human prior knowledge to generate better prompts?}

Driven by them, we propose a novel evidential prompt generation method named \textbf{\mName}tailored for SAM in medical images. Notably, to minimize the need for labels and computational resources to the greatest extent possible, \mName is \textbf{training-free} and requires only a \textbf{single} medical image-annotation pair as a reference. Addressing question Q1, we propose an evidential learning method to propagate the single annotation to other images. Specifically, we propose to stitch the annotated reference image and unlabeled images for SAM to extract features. By estimating the similarities between anchor features (from ground-truth ROIs) and others as evidence, we propose to select the points with the highest belief mass as prompt points. For question Q2, we turn to two straightforward common sense as prior knowledge. \textbf{First}, semantic is invariant to photometric transformations and equivariant to geometric transformations. \textbf{Second}, medical images of the same modality and category often share similar appearances due to the constraints of anatomical structures, which in contrast with natural images that typically exhibit rich diversity and encompass a wide range of classes. Based on the first prior, we generate a perturbation set for each target image and combine their belief mass. With the latter, we spatially assemble the reference perturbation set and target images to embed them into the same feature space. In this way, the rich contextual information is extracted by the model's comprehensive attention module to guide the generation of prompts. 

By answering two questions, \mName presents a robust and efficient approach to medical image segmentation using SAM. To summarize, the major contributions of our work are as follows:
\begin{itemize}
    \item We introduce an evidential method based on uncertainty estimation which automatically generates prompts for SAM in medical images without the interaction of clinical experts. 
    
    \item We incorporate the human prior into the prompts. This step is vital for alleviating the domain gap between natural and medical images and enhancing the applicability and usefulness of SAM in medical scenarios.

    \item We evaluate our method on a broad range of tasks and modalities without training and only one reference annotation.
\end{itemize}

It's important to note that the primary objective of this work is not to achieve new state-of-the-art results or to surpass existing methods across all benchmarks. Instead, we focus on offering a simple and efficient approach to harness foundational computer vision models.

 
\section{Related Work}
\textbf{Image Segmentation.} Image segmentation aims to perform pixel-level classification. Depending on specific focus areas, segmentation is categorized into semantic segmentation (involving predefined classes) \cite{zhou2022rethinking, guo2022segnext}, instance segmentation (focused on the identification of individual objects) \cite{huang2022minvis, ke2022mask}, and panoptic segmentation (which encompasses both semantic classification and object identification) \cite{li2022panoptic, mohan2022amodal}. Beyond its applications in natural images, segmentation of organs or lesions in medical images plays a critical role in computer-aided clinical diagnosis \cite{su2023slaug, you2022class}. Recently, the Segment Anything Model (SAM) \cite{kirillov2023segany} introduced a promptable segmentation task, achieving state-of-the-art performance. However, when applied to medical images, it necessitates substantial human prior knowledge to achieve appealing results \cite{ji2023segment}. In this work, we present a novel method designed to automatically generate point prompts for medical images, thereby reducing the need for human interaction.

\noindent \textbf{Evidential Learning and Uncertainty.} Given that deep models often exhibit undue confidence in incorrect predictions \cite{amodei2016concrete}, the estimation of uncertainty becomes essential for reliable decision-making, especially in high-risk fields such as clinical applications. Assembling neural networks represents a straightforward and scalable approach to uncertainty estimation, with proven success in enhancing performance \cite{lakshminarayanan2017simple}. The Dempster-Shafer theory has been utilized to estimate uncertainty across multiple views for classification \cite{han2022trusted} and has found applications in medical image segmentation \cite{gao2023reliable}. While these methods have been effective in network optimization, the uncertainty within large vision foundation models remains an unexplored area. Our experimental findings demonstrate that our proposed method effectively gauges uncertainty during the reference process of large models, boosting performance without the necessity for additional training.

\noindent \textbf{In-Context Visual Learning.} Originally, in-context learning is a paradigm first introduced in GPT-3 \cite{NEURIPS2020_1457c0d6} that empowers language models to learn specific tasks with only a few examples \cite{dong2022survey}. Extending this concept to computer vision, some researches \cite{bar2022visual, wang2023images} concatenate the image with its target into a single entity and then apply standard masked image modeling to learn in-context information. Building on this framework, \cite{wang2023seggpt} unifies various segmentation tasks and devises a novel random coloring scheme in in-context training. Diverging from existing approaches that rely on inpainting tasks, our work captures the in-context information among the ensembled reference and target images directly into the inference stage. Additionally, we present a training-free method specifically designed to harness the capabilities of SAM in the field of medical image segmentation.

\begin{figure*}[t]
    \centering
    \includegraphics[width=0.95\linewidth]{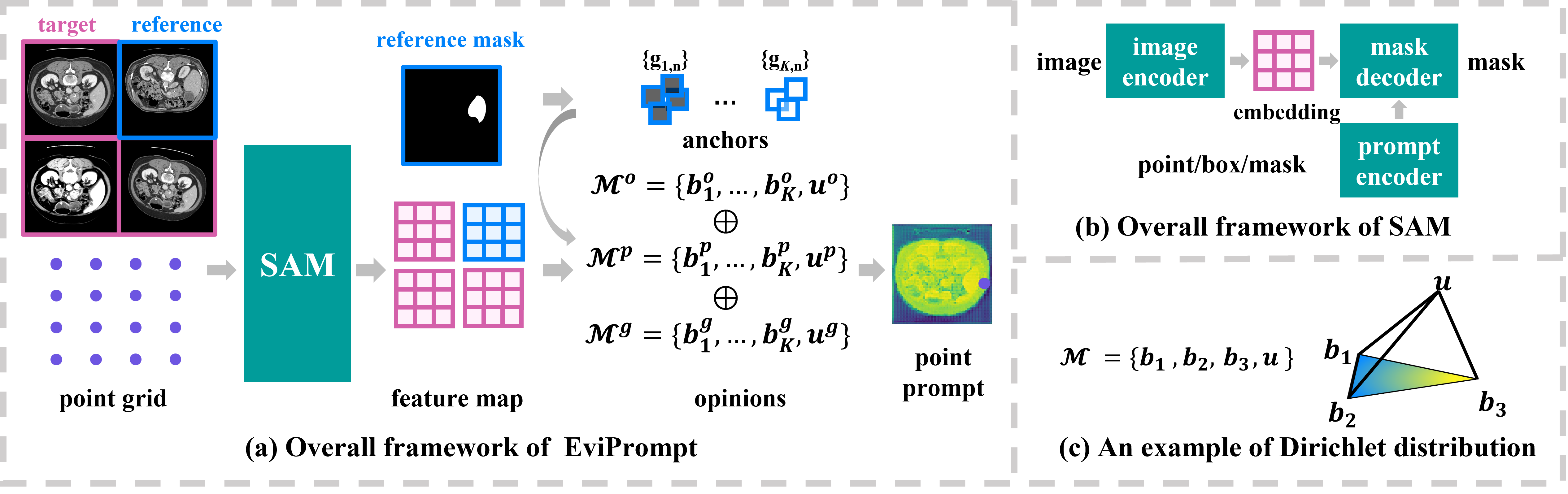}
    \caption{(a) Overall framework of EviPrompt, which is training-free and only needs single medical image-annotation pair as reference. We extend the target image into a perturbation set and spatially ensemble them with reference images as input of SAM to explore the rich contextual information. Then, the reference mask and the feature map are utilized to generate anchors for the belief to select point prompts. (b) Overall framework of SAM, is composed of three components: an image encoder, a flexible prompt encoder, and a lightweight mask decoder. (c) An example of Dirichlet distribution was introduced to estimate the uncertainty. Based on the subjective logic, the Dirichlet distribution is converted into a standard 3-simplex, and the opinion $\mathcal{M}=\{b_1,b_2,b_3, u\}$ is the point in the simplex. }
    \label{fig:framework}
\end{figure*}

\noindent \textbf{Foundation Models.} Foundation models exhibit impressive performance across a wide array of downstream tasks, demonstrating remarkable generalization capabilities. In the field of natural language processing (NLP), models such as BERT \cite{devlin2018bert}, GPT \cite{NEURIPS2020_1457c0d6}, and LLaMA \cite{touvron2023llama} can be seamlessly applied to new language tasks, using carefully crafted prompts at inference. When fine-tuned on medical data, these foundation models have also shown significant advancements in understanding patient inquiries and offering precise guidance \cite{yunxiang2023chatdoctor, singhal2023large, singhal2023towards}. In computer vision, CLIP \cite{radford2021learning} transfers the success of task-agnostic pre-training from NLP, utilizing image-text pairs for supervision and achieving results on par with fully supervised methods. Another notable model is SAM \cite{kirillov2023segany}, which is trained on 1 billion masks and specifically designed for image segmentation. Recent studies \cite{wang2023mathrm, li2023polyp, MedSAM} indicate that fine-tuning SAM on medical images substantially improves segmentation accuracy for clinical imaging applications. Our approach differs by adopting a more efficient method that does not require external data or additional training. 


\section{Method}

\subsection{Overall Framework} 
Despite SAM's powerful zero-shot transfer capability, two challenges lie in its application to medical images: the requirement of human labor and the domain gap between natural and medical images. To this end, We build a \textit{training-free} evidential prompt generation method named \textit{EviPrompt}. The overall framework is illustrated in Fig.\ref{fig:framework}(a). Our goal is to segment the target data $\{\textbf{x}_t\}$ with only a \textit{single} medical image-annotation pair, denoted as $(\textbf{x}_r, \textbf{y}_r)$. Without any training procedure, the overall strategy is two-folded, where we first compare the annotated data with all the other images and then schedule the annotation propagation manner and generate the synthetic prompts for mask prediction. For brevity, we term the single image-annotation pair as the reference, while all the other images under segmentation as the target.

Given a very limited but practical scenario of single image-annotation training pair, we propose a novel sub-module that \textit{explicitly compare} the reference image and the target image under segmentation. Instead of encoding the reference and target features separately, we designed a 2x2 comparator board, stitching the reference and other 3 target images as the original input to the SAM model (see Fig.\ref{fig:framework} for details). The advantage of this is to provide better contextual modeling. Once the stitched features are obtained via SAM, we calculate the similarities between the target features and the anchor features (corresponding to the single segmentation annotation). The similarity maps are ready to generate the prompts. 

The stitched similarity maps can be divided into multiple folds corresponding to different augmentations on the target image. Taking a single target image as an example, each fold of the similarity map represents \textit{how the expert views a specific kind of augmentation on the target and all folds foster a multi-expert committee with different ``subjective'' opinions}. The SAC adopts multiple augmentations inspired by the human prior knowledge (in the medical annotation procedure) and proposes to estimate a comprehensive similarity map based on evidential learning, by combining the uncertainties of each similarity, mimicking the decision-making process in the annotation committee. The potential advantages over straight-forward fusion are to avoid the over-confidence estimation as shown in Fig.\ref{fig:teaser}(d).





\subsection{Preliminary}
\textbf{Revisit of SAM.} SAM introduce a novel task, \textit{promptable segmentation}, which is to generate mask with prompt. As illustrated in Fig.\ref{fig:framework}(b), the model consists of three components: an image encoder, a flexible prompt encoder, and a lightweight mask decoder. SAM takes an image and a set of prompts (points, boxes, and masks) as input. Then, the image embedding and encoded prompts are fed into the mask decoder for generating the final mask prediction with the attention mechanism.

\noindent \textbf{Uncertainty Estimation.} 
Subjective logic \cite{jsang2018subjective} is an uncertain probabilistic logic to address formal representations of trust. Specifically, considering the $K$ classification problems, given the input sample, $\textbf{e}=[e_1,...,e_K]$ with non-negative elements is the \textit{evidence} which refers to the metrics to support classification. Then, the \textit{belief mass} of the $k$-th class $b_k$, and the overall \textit{uncertainty} $u$ is computed using the evidence:
\begin{align}
\label{evi}
    b_k = \frac{e_k}{S} \text{ and } u= \frac{K}{S},
\end{align}
where $S=\sum_{k=1}^K (e_k+1)$ is the Dirichlet strength. Thereby, we have an \textit{opinion} $\mathcal{M}=\{b_1,...,b_K, u\}$, satisfying $\sum_{i=1}^K b_k+u=1$. In subjective logic, the class probabilities, which correspond to opinion, are represented by Dirichlet distribution $Dir(\textbf{p}|\textbf{a})$ where $a_k=e_k+1$. Intuitively, we illustrate an example of 3-classes classification in Fig.\ref{fig:framework}(c). The expected probability for the $k$-th class is the mean of Dirichlet distribution and obtained as $\hat{p}_k=\frac{a_k}{S}$ \cite{Frigyik2010IntroductionTT}. 

\subsection{Human Prior Knowledge }
In this section, we elucidate the human prior knowledge that directs the generation of prompts to alleviate the domain gap between the training data (i.e., natural images) and the target data (i.e., medical images). Although variations in scanners and scanning parameters can alter the appearance of images, these modifications don't impact the subject's actual physical condition, \ie the anatomical structure. We characterize these variations in image appearance as photometric and geometric transformations. Hence, we have prior knowledge\ding{182}: \textit{semantics remain invariant to photometric transformations and are equivariant to geometric transformations}. Concurrently, the uniformity of anatomical structures results in a consistent appearance across medical images. This inherent characteristic gives rise to our prior knowledge\ding{183}: \textit{pixels/voxels within the same category exhibit more similar appearances compared to those from different categories}.

\subsection{Prompt Generation} 
Informed by the prior knowledge\ding{182}, we transform each target image into a perturbation set, denoted as ${\textbf{x}_t^o, \textbf{x}_t^g, \textbf{x}_t^p}$. Here, $\textbf{x}_t^o=\textbf{x}_t$ represents the original image, $\textbf{x}_t^g = T^g(\textbf{x}_t)$ represents its geometrically transformed version (e.g., scaling, skewing, rotation), and $\textbf{x}_t^p = T^p(\textbf{x}_t)$ designates the photometrically transformed version (e.g., color jittering). Unlike natural images that exhibit diverse classes across different images, medical images typically contain the same target class. We capitalize on this characteristic by arranging the perturbation set and reference image within a $2\times2$ grid, ensuring both the reference and target images are embedded in a shared feature space. This arrangement encourages the model to extract rich contextual information through its comprehensive attention module. To steer the model's focus evenly across the image and discover all potential proposals, we propose the point grid as the prompt. 

Guided by the prior knowledge\ding{183}, we construct prompts based on pixel similarities. However, even pixels belonging to the same category can exhibit varying appearances. A quintessential example is the segmentation of the liver in MR imaging. Here, the background isn't merely an undifferentiated mass; it comprises diverse structures such as the spleen, aorta, and right kidney. For example, in the liver segmentation task in CT, the background comprises diverse structures such as the spleen, aorta, and right kidney. Given this inherent complexity and diversity, we posit that a single anchor representation may not suffice to adequately capture the rich physiological nuances of each class. As an alternative, we propose a collection of anchors to represent each category. Specifically, we utilize the output before the classification head of the mask decoder as the feature map and subsequently reverse the spatial assembly into $\textbf{F}_r$, $\textbf{F}_t^o$, $\textbf{F}_t^g$, and $\textbf{F}_t^p$, which correspond to $x_r$, $x_t^o$, $x_t^g$, and $x_t^p$ respectively. We perform clustering for the features in $\textbf{F}_r$ of each class to generate centroids serving as anchors, denoted as $\{\textbf{g}_{k,n}\}_{n=1}^{N_k}$. Given pixel $i$, the maximum value of the similarity between the anchor set and the pixel-level features, $\textbf{F}[i]$, represents the pixel-level evidence:
\begin{align}
\label{anchor}
e_k[i] = \max_n \sigma(\textbf{F}[i]^\top \textbf{g}_{k,n}),
\end{align}
where $\sigma(\cdot)$ represents an activation function (e.g., SoftPlus) to ensure the evidence remains non-negative. 


By applying Eq.\ref{anchor} and \ref{evi}, we derive three opinions for each target image from the perturbation set: $\mathcal{M}^o$, $\mathcal{M}^p$, and $\mathcal{M}^g$. These opinions are consequent from $\textbf{e}^o$, $\textbf{e}^p$, and ${T^g}^{-1} (\textbf{e}^g)$ respectively. With these opinions in place, we then combine them to obtain an overall opinion which can subsequently be translated into the distribution guiding the final prediction.

\noindent \textbf{Dempster’s combination rule} \cite{josang2012interpretation} \textit{Given opinions $\mathcal{M}^1=\{b_1^1,...,b_K^1, u^1\}$ and $\mathcal{M}^2=\{b_1^2,...,b_K^2, u^2\}$, the combined opinion $\mathcal{M}=\{b_1,...,b_K, u\} = \mathcal{M}^1 \oplus \mathcal{M}^2$ is obtained as follows}:
\begin{align}
    b_k = \frac{1}{1-C}(b_k^1b_k^2+b_k^1u^2+b_k^2u^1), u= \frac{1}{1-C} u^1u^2,
\end{align}
\textit{where $C=\sum_{i\neq j} b_i^1b_j^2$ signifies the conflict between two opinions.}

Subsequently, the final opinion is obtained by applying the above rule 3 times $\mathcal{M}=\{b_1,...,b_K, u\} = \mathcal{M}^o \oplus \mathcal{M}^p \oplus \mathcal{M}^g$. 


\subsection{Mask Prediction}
\noindent \textbf{Point Selection.} SAM exhibits sensitivity to edges \cite{kirillov2023segany}, often leading to an exaggerated belief mass at these boundaries. When choosing points based solely on the highest $b_k$ values, there's a strong tendency to select those located on the edges, which potentially results in ambiguous predictions. To address this, we apply averaging filtering in the spatial domain. Post this correction for edge prominence, another challenge emerges – the innate similarity among neighboring points. This causes the direct point selection strategy to often yield closely situated points. To overcome this limitation, we introduce a patch-based point selection method. We spatially divide the feature map into square patches of dimensions $p \times p$. Subsequently, we aggregate the belief values inside each patch and proceed to select the most possible patches. Within each of these patches, the point with the highest belief mass is chosen to serve as the point prompt. (see Section Analysis for more discussion).

\noindent \textbf{Mask Refinement.}
Once SAM takes the image and point prompts, it produces a predicted mask. The refinement phase follows this prediction. We transform the mask into a bounding box, defined by its maximum and minimum positional extents. This box, in conjunction with the point prompts, acts as the input for the second mask generation. Leveraging shared image embeddings ensures only a slight increase in the computational demand of the prompt encoding process, making this refinement step computationally efficient.

\section{Experiments}
\subsection{Experiment Setup}
\begin{figure}[t]
    \centering
    \includegraphics[width=0.85\linewidth]{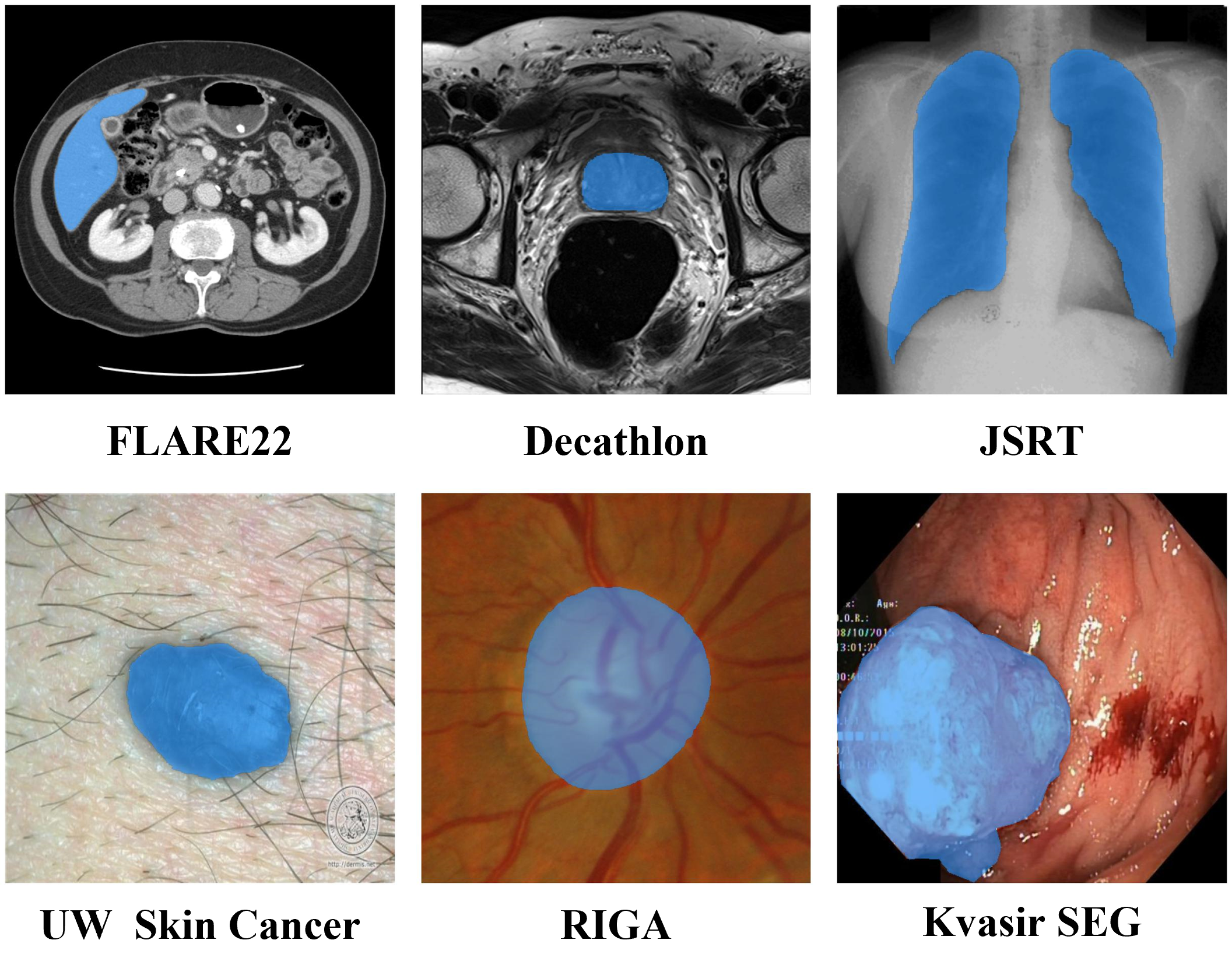}
    \caption{Examples from the utilized dataset. The highlighted zones are the anatomy of interest}
    \label{fig:dataset}
\end{figure}

\begin{table}[t]
\centering
\small
\begin{tabular}{lcccc}
\toprule
Dataset& Modality& Object Class& \#Samples\\
\midrule
FLARE22 &CT &Liver &50\\
Decathlon  &MR &Prostate &32\\
JSRT & CXR& Lung& 60\\
UW Skin Cancer &  Dermoscopy& Skin cancer&  206\\
RIGA & Fundus& Retinal disc& 750\\
Kvasir SEG&  Endoscopy &Polyp &1000\\






 
\bottomrule
\end{tabular}
\caption{A summary of datasets used for evaluations.}
\label{tab:dataset}
\end{table}
\textbf{Datasets.} Extensive experiments are conducted to verify the effectiveness of the proposed framework on six medical segmentation tasks from varied image modalities: FLARE22 \cite{ma_2023_7860267}, FLARE22 \cite{ma_2023_7860267}, JSRT \cite{shiraishi2000development}, UWaterloo Skin Cancer \cite{12}, RIGA \cite{almazroa2017agreement} and Kvasir SEG \cite{jha2020kvasir}. A brief description of these datasets is summarized in Table \ref{tab:dataset}, and examples are shown in Fig.\ref{fig:dataset}. More details can be found in Supplementary.

\noindent \textbf{Implementation details.} 
In all experiments, we do \textbf{not} update the weights of the model. For geometrical transformation, scaling, skewing, and rotation are implemented. For photometric transformation contrast and color saturation are altered. To comply with the model's input prerequisites, following \cite{MedSAM}, the intensity values of CT, and MR are rescaled to fall within the range of $[0,255]$ and replicated three times along the channel axis. All images are resized to a uniform dimension of $512 \times 512 \times 3$. These experiments are conducted using PyTorch and executed on a Nvidia 3090 GPU. More details are in the supplementary.

\noindent \textbf{Evaluation Metrics.}
We propose two widely-used metrics for medical image segmentation: the Dice Similarity Coefficient (DSC) and the Normalized Surface Distance (NSD). The DSC is employed to evaluate the region overlap between the ground truth and the segmented results, providing a measure of the similarity between the two areas. And the NSD assesses the consensus between the boundaries of the ground truth and segmentation results at a given tolerance level. 

\begin{figure}[t]
    \centering
    \includegraphics[width=0.95\linewidth]{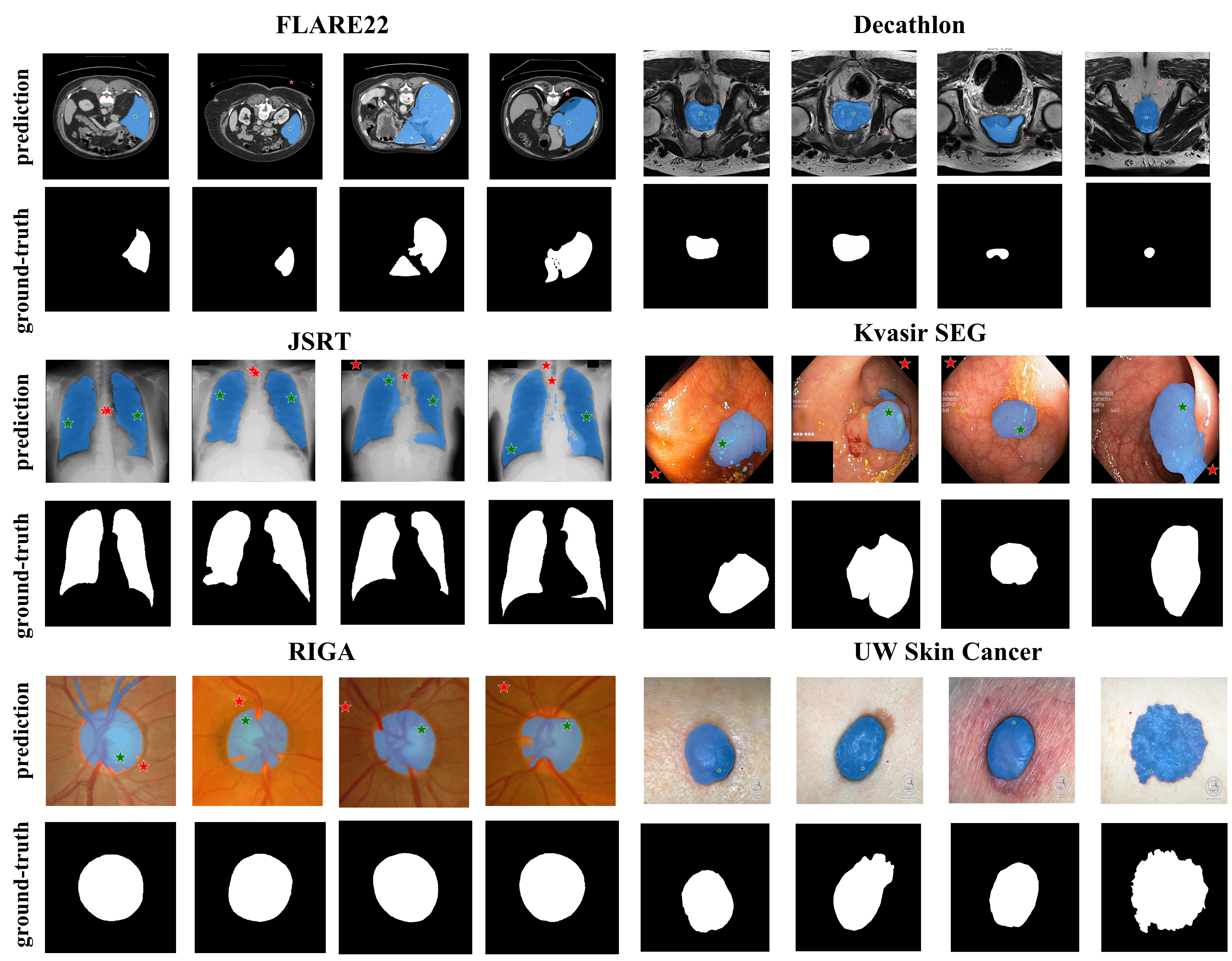}
    \caption{Analysis of the success and failure cases }
    \label{fig:cas}
\end{figure}

\subsection{Results}
In this section, we perform segmentation in the prompt mode with points. We conduct experiments with SAM with varying numbers of points and backbone networks. Additionally, besides the official weight, we also investigate the MedSAM \cite{MedSAM} which is fine-tuned on SAM with image datasets. The results are reported in Table \ref{tab:result}. To investigate the capability of SAM and our method from an intuitive perspective, we visualize the success and failure case analysis shown in supplementary materials due to space limits.

\begin{table*}[t]
\centering
\small
\resizebox{0.95\linewidth}{!}{
\begin{tabular}{lllcccccccccccccc}
\toprule
\multirow{2}{*}{\#Points} &\multirow{2}{*}{Weight} & \multirow{2}{*}{Method} & \multicolumn{2}{c}{FLARE22} & \multicolumn{2}{c}{Decathlon}& \multicolumn{2}{c}{JSRT}& \multicolumn{2}{c}{UW Skin Cancer}&  \multicolumn{2}{c}{RIGA}& \multicolumn{2}{c}{Kvasir SEG}\\
& && DSC$\uparrow$ & NSD$\uparrow$ & DSC$\uparrow$ & NSD$\uparrow$ & DSC$\uparrow$ & NSD$\uparrow$
& DSC$\uparrow$ & NSD$\uparrow$ & DSC$\uparrow$ & NSD$\uparrow$ & DSC$\uparrow$ & NSD$\uparrow$\\
\midrule
\multicolumn{15}{c}{Image encoder: ViT-B}\\
\midrule
\multirow{4}{*}{1} & \multirow{2}{*}{SAM}& Oracle & 63.97& 39.53& 59.24& 15.56& 65.10& 9.38& 72.53& 41.98&35.93& 2.91&56.76& 22.45&\\
& & EviPrompt& 44.33& 17.51& 46.20& 7.69&64.06 &7.13 & \textbf{76.39}& 27.93& \textbf{40.64}&\textbf{3.08}&36.10&8.00\\
& \multirow{2}{*}{MedSAM}& Oracle& 6.99& 4.80&  8.64& 3.95& 19.82& 4.68& 16.48&  7.21&12.44& 3.29&11.63& 3.66\\
& & EviPrompt&\textbf{10.92} &4.76 & \textbf{16.96}& 2.20&\textbf{25.15 }&3.24 &\textbf{29.75} &3.16& \textbf{23.64}& 2.26&\textbf{14.13}&2.12\\
\midrule
\multirow{4}{*}{3} & \multirow{2}{*}{SAM}& Oracle & 73.99& 44.72&59.60& 14.60& 77.49& 14.12& 84.77& 44.94&55.76&5.98&74.27& 25.00\\
& & EviPrompt&44.18 & 13.88& 44.36& 6.10&\textbf{79.67} &11.60 & 80.90& 28.20&\textbf{63.73}&5.21&42.62&9.25 \\
& \multirow{2}{*}{MedSAM}& Oracle& 9.05& 3.83& 12.35& 3.26&38.31& 5.73& 32.24&  9.17&21.96& 3.92&16.58& 3.75\\
& & EviPrompt&\textbf{20.20} &\textbf{5.42} &\textbf{22.37}& 1.67& \textbf{59.44}& \textbf{5.80}& \textbf{48.30}&7.57&\textbf{50.21}&3.84&\textbf{24.45}&2.65\\
\midrule
\multirow{4}{*}{5} & \multirow{2}{*}{SAM}& Oracle & 71.63& 39.38& 52.54& 9.91& 78.31& 15.59& 85.94& 45.75&57.70&5.95&74.62& 24.03\\
& & EviPrompt& 36.04& 7.02& 40.15&4.68 &\textbf{79.75} &12.22 & 77.14&24.69&\textbf{65.75}&5.20&42.09&8.58\\
& \multirow{2}{*}{MedSAM}& Oracle& 6.76& 2.26& 9.93& 2.34& 49.12& 7.38& 39.85&  8.25&26.81& 3.87&18.11&3.40\\
& & EviPrompt& \textbf{19.26}& \textbf{4.33}& \textbf{19.77}& 0.94&\textbf{67.38}&\textbf{7.64} & \textbf{52.05}& 7.41& \textbf{59.15}& \textbf{4.43}&\textbf{28.42}&2.55 \\
\midrule

\multicolumn{15}{c}{Image encoder: ViT-H}\\
\midrule
\multirow{2}{*}{1} & \multirow{2}{*}{SAM}& Oracle & 68.48& 45.07& 60.45& 16.08& 55.90& 3.81& 79.03& 45.87&58.40& 8.39& 64.34& 27.56
\\
& & EviPrompt &38.18&15.52 & 46.13& 7.23&55.23 &2.91 & \textbf{80.67}& 28.77& \textbf{58.42}&6.31&42.13&11.01\\
\midrule
\multirow{2}{*}{3} & \multirow{2}{*}{SAM}& Oracle & 70.11& 44.88& 59.34& 15.11& 73.07& 13.09& 86.73& 47.17&69.07&10.04&77.86& 30.59\\
& & EviPrompt& 36.99&10.50 & 39.84&5.22 &\textbf{80.28} &12.74 & 81.16& 27.74&\textbf{69.37}&7.41&47.83&11.41\\

\midrule
\multirow{2}{*}{5} & \multirow{2}{*}{SAM}& Oracle & 68.81& 41.44& 52.14& 11.38& 73.34& 13.25&88.05& 45.80&72.86&10.20&80.83& 30.06\\
& & EviPrompt& 37.88& 9.19& 34.48& 3.45&\textbf{88.27} &\textbf{15.83} &76.42 &24.12 &\textbf{74.14}&8.20&49.82&11.23\\
\bottomrule
\end{tabular}}
\caption{Performance comparison on 6 datasets.}
\label{tab:result}
\end{table*}
\noindent \textbf{Performance of SAM and MedSAM.}
We introduce an "oracle" assessment where point prompts are derived uniformly from the ground truth mask, simulating an idealized form of human interaction. The results are fascinating: despite being primarily trained on natural images, SAM showcases impressive prowess when guided by these "perfect" prompts. Broadly, SAM excels in scenarios with large target areas (\eg liver) or where there is a pronounced color contrast between the object and the background, as seen in the case of skin cancer. However, challenges arise when the model confronts scenarios where the target object is diminutive (\eg prostate) or the color closely matches the background (\eg polyps). MedSAM surprisingly trails behind in certain tests. We hypothesize that the root of this inconsistency lies in its training approach. Specifically, MedSAM's training largely relies on bounding boxes as prompts rather than the points in our experiments. A striking feature of SAM's performance is its variability. In some instances, it produces near-flawless segmentations, seamlessly segmenting the target object from its surroundings. In contrast, there are instances where it completely fails to recognize the target, leaving it undetected. 

\noindent \textbf{Performance of SAM / MedSAM + EviPrompt.}
Our proposed approach demonstrates commendable results across various tasks, often very competitive ones, and in certain datasets like RIGA, even outperforming SAM + oracle. The superior performance in RIGA can be attributed to the precision of our prompts and the subsequent mask refinement steps. A significant strength of our methodology emerges in the segmentation of anatomical structures. The inherent consistency in the appearance of these structures across different samples ensures that our reference-guided approach can harness rich contextual cues, facilitating more accurate prompt generation. Our approach shows promising results in medical image segmentation scenarios. In addition, our methodology presents a substantial enhancement over MedSAM in the point prompts mode. The outperformance over MedSAM not only demonstrates the effectiveness of our approach but also conveys wide usage of SAM for medical image applications using very limited resources and annotations. The amalgamation of SAM's inherent capabilities with the reduction in manual interventions of our prompt generation can pave the way for more automated, yet accurate, medical imaging analyses.

\begin{figure}[t]
    \centering
    \includegraphics[width=0.95\linewidth]{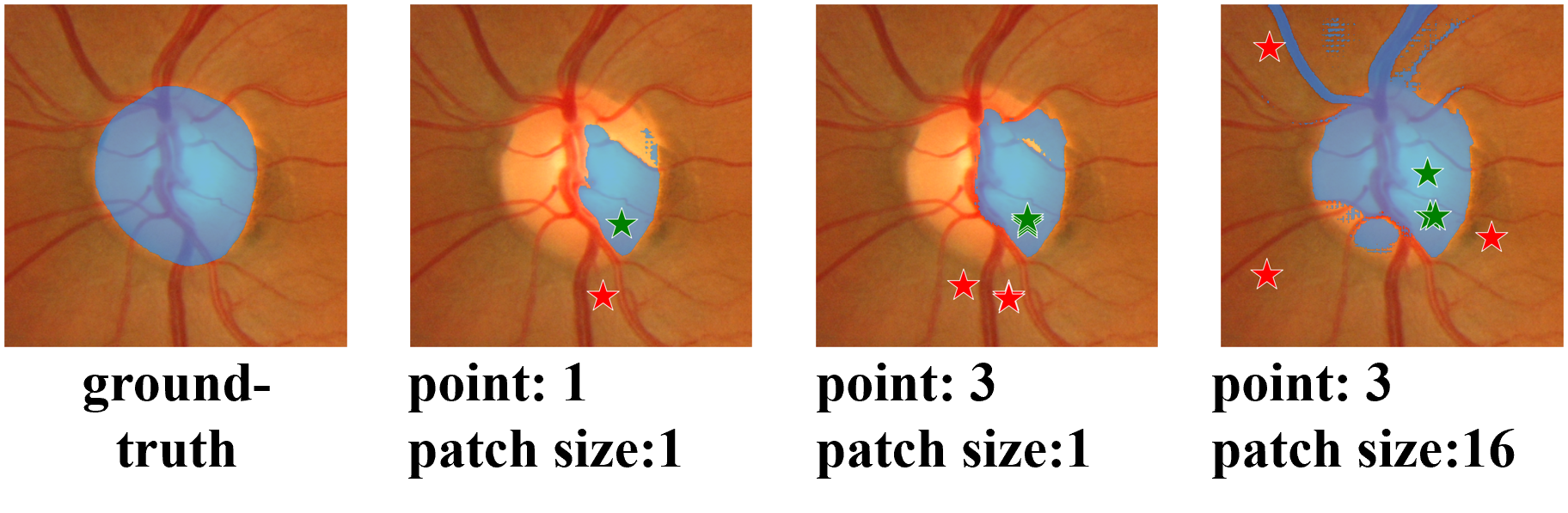}
    \caption{The chosen points tend to be close to each other (in the aspect of positions) With smaller patch sizes, making their collective impact akin to that of a single point.}
    \label{fig:patchsize}
\end{figure}

\subsection{Analysis}
\label{Analysis}
In this section, we investigate the effect of our core designs. We conduct experiments using SAM as the model weight with ViT-b as the image encoder.
\begin{table}[t]
\centering
\small
\resizebox{0.95\linewidth}{!}{
\begin{tabular}{lcccccc}
\toprule
Patch Size& 1& 4& 8& 16& 32& 64\\
\midrule
DSC&52.24&55.65&58.95& 63.73&63.33&62.83\\
NSD&4.45&4.55& 4.73&5.21&5.11&5.28\\
\bottomrule
\end{tabular}}
\caption{Analysis of patch size on RIGA.}
\label{tab:patch size}
\end{table}
\begin{table}[t]
\centering
\small
\begin{tabular}{lcccccc}
\toprule
$N_0$& 1& 5& 5& 10& 10&\\
$N_1$& 1& 1& 5& 5& 10& \\
\midrule
DSC&40.41& 42.95& 42.95&44.36&44.36\\
NSD&5.83& 5.60& 5.60& 6.09&6.10\\
\bottomrule
\end{tabular}
\caption{Analysis of the number of anchors on Decathlon.}
\label{tab:anchors}
\end{table}

\noindent \textbf{Number of Points.} In natural image segmentation tasks, the performance of SAM improves with the increasing number of point prompts\cite{kirillov2023segany}. The trend is also seen in medical images as per Table \ref{tab:result}. However, the performance of the proposed method degrades when adding more points for some tasks. This is because increasing points based on belief mass also increase the risk of selecting less accurate points, especially for small anatomical structures under segmentation. Hence, there's a trade-off: achieving optimal performance requires balancing the quantity with the quality of points.

\noindent \textbf{Patch Size $p$.} 
Table \ref{tab:patch size} reports the performance when the patch size varies from 1 to 64 on RIGA. The result demonstrates a significant influence of patch size on the final performance. For $p=1$, points are directly selected from the feature map, resulting in a DSC of $52.24\%$. With smaller patch sizes, the chosen points tend to be close to each other (in the aspect of positions), making their collective impact akin to that of a single point, as shown in Fig \ref{tab:patch size}. However, with an increased patch size (\ie $p=16$), there's a notable improvement in performance (\ie $52.24\%\to63.73\%$). Larger patch sizes encourage the selection of points that are farther apart from each other. Further expanding the patch size leads to performance stabilization.

\noindent \textbf{Number of Anchors.}
Table \ref{tab:anchors} reports the performance on Decathlon with regard to the number of anchors for background and foreground, denoted as $N_0$ and $N_1$, respectively. When both $N_0$ and $N_1$ are set to 1, the class is represented by the mean of its pixel-level features, resulting in a DSC of $40.41\%$. Notably, expanding the number of anchors to 5 for both background and foreground ($N_0=N_1=5$) enhances the performance, elevating the DSC to 
$42.95\%$. The metric further advances with 10 anchors for each class. We observe that the rise in $N_0$ yields more pronounced gains than equivalent increments in $N_1$. This can be attributed to the background's diverse appearance relative to the foreground.

\begin{table}[t]
\centering
\small
\begin{tabular}{lcccccc}
\toprule
Variant & DSC & NSD \\
\midrule
Eviprompt& 63.73&5.21\\
\midrule
w/o transformations& 59.12& 5.08\\
w/o evidential learning& 58.16& 4.85\\
w/o mask refinement& 63.51& 5.12\\
\bottomrule
\end{tabular}
\caption{Ablation Study on RIGA with 3 points.}
\label{tab:ablation}
\end{table}

\noindent \textbf{Ablation Study.}
To investigate the effect of different components in our method, we conduct the ablation study on RIGA: (1) w/o transformations: we replace the transformed version ($x_t^g$ and $x_t^p$) with the original image $x_t$. (2) w/o evidential learning: we replace the belief mass with the probability derived from the softmax function and replace the opinion combination with the mean value. (3) w/o mask refinement. The results are shown in Table \ref{tab:ablation}. We observe that the full method outperforms other variants. w/o evidential learning suffers an obvious degradation of $8.15\%$ of DSC. And removing transformations and mask refinement degrades the DSC to $59.12\%$ and $63.51\%$, which verified their effectiveness.

\section{Discussion}
We have shown the SAM's capability on medical images across a variety of modalities and tasks. However, its performance still lags behind some models trained on specialist data. Particularly, its struggles in scenarios without a distinct color differentiation between the foreground and background. As detailed in the experiments section, evaluating SAM's worth should not merely rest on its comparative performance. These observations emphasize the inherent challenges and opportunities in adapting models, primarily trained on generic datasets, to the specialized domain of medical imaging. The uneven performance of SAM reiterates the importance of domain-specific training. Nevertheless, the glimpses of excellence seen in SAM's "oracle" assessments offer hope and hint at the vast potential lying dormant, waiting to be harnessed with the right tweaks and strategies. We hope that our work will inspire the community to delve deeper into harnessing the might of foundational natural image models effectively in medical images. 

\section{Conclusions}

In this paper, we propose a novel systematic solution termed as EviPrompt, that is training-free and requires single medical image annotation. EviPrompt made several technical contributions including a novel setup to extract multi-fold features in SAM and the evidential learning guided automatic prompt generation. EviPrompt has been evaluated on a broad range of tasks and modalities in medical image segmentation and achieves the SOTA results.

{
    \small
    \bibliographystyle{ieeenat_fullname}
    \bibliography{main}
}


\end{document}


\maketitle
\begin{figure*}[t]
    \centering
    \includegraphics[width=0.8\linewidth]{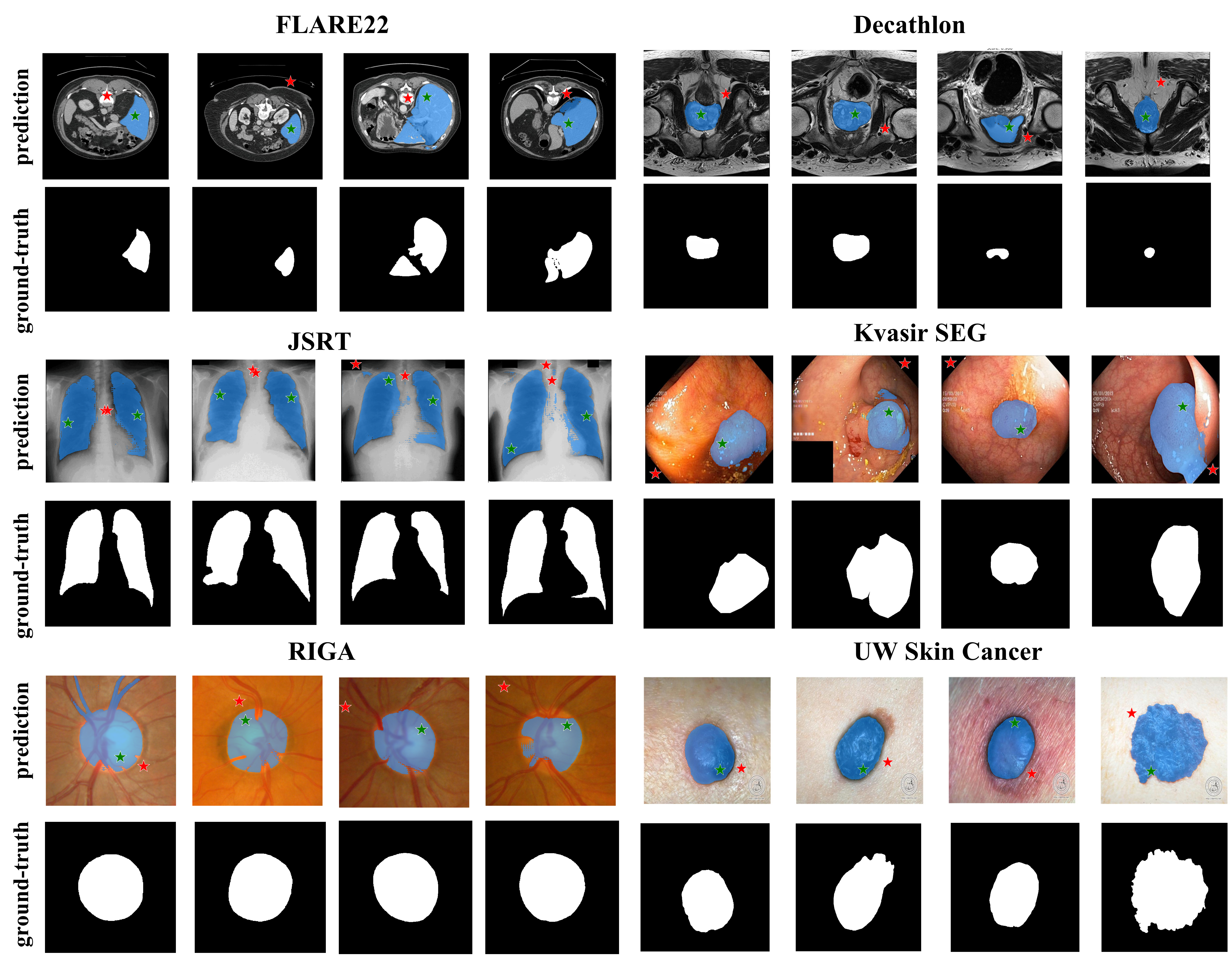}
    \caption{More results.}
    \label{fig:res}
\end{figure*}

\section{Experiment Setup}
\subsection{Datasets}
\noindent \textbf{FLARE22} \cite{ma_2023_7860267} is curated from more than 20 medical groups under the license permission. In this paper, we utilize the training set which includes 50 labeled CT scans with liver, kidney, spleen, or pancreas diseases.

\noindent \textbf{Decathlon} \cite{Antonelli_2022} consists of 48 prostate multiparametric MRI (mpMRI) studies comprising T2-weighted, Diffusion-weighted and T1-weighted contrast enhanced series. A subset of two series, transverse T2-weighted and the apparent diffusion coefficient (ADC) was selected. The corresponding target ROIs were the prostate peripheral zone (PZ) and the transition zone (TZ). We utilize the T2 series of the labeled training set.

\noindent \textbf{JSRT} \cite{shiraishi2000development}. contains 60 images from the "Standard Digital Image Database [Chest Mass Shadow Image]" of the JSRT. The aim is to extract the lung field area on chest X-ray image.

\noindent \textbf{UWaterloo Skin Cancer} \cite{12} contains 206 images of skin lesions including 119 melanomas, and 87 not melanoma. They are obtained using standard consumer-grade cameras in varying and unconstrained environmental conditions from the online public databases Dermatology Information System and DermQuest.

\noindent \textbf{RIGA} \cite{almazroa2017agreement} contains in total of 750 color fundus images from three sources, including 460 images from MESSIDOR, 195 images from BinRushed
and 95 images from Magrabia. All images are labeled by six glaucoma experts from
different organizations, and we utilize the final calibrated segmentation maps as the ground-truth.

\noindent \textbf{Kvasir SEG} \cite{jha2020kvasir} contains 1000
annotated polyp images and their corresponding masks. 

\subsection{Implementation details.} 
In all experiments, we do \textbf{not} update the weights of the model. For geometrical transformation, scaling, skewing, and rotation are implemented. For photometric transformation contrast and color saturation are altered. To comply with the model's input prerequisites, following \cite{MedSAM}, the intensity values of CT, and MR are rescaled to fall within the range of $[0,255]$ and replicated three times along the channel axis. In the prompt generation process for the JSRT dataset, each image is partitioned vertically into two sections, with prompts generated for both. For all experiments, we consistently utilized a patch size of 16.

\section{More results}
Fig.\ref{fig:res} provides more results of our method in the six datasets. 

\bibliography{ref}